\documentclass[sn-mathphys]{sn-jnl}% Math and Physical Sciences Reference Style
\usepackage{graphicx}%
\usepackage{multirow}%
\usepackage{amsmath,amssymb,amsfonts}%
\usepackage{amsthm}%
\usepackage{mathrsfs}%
\usepackage[title]{appendix}%
\usepackage{xcolor}%
\usepackage{textcomp}%
\usepackage{manyfoot}%
\usepackage{booktabs}%
\usepackage{algorithm}%
\usepackage{algorithmicx}%
\usepackage{algpseudocode}%
\usepackage{listings}%
\usepackage{caption}

\usepackage{wasysym}
\usepackage{array}

\usepackage{textcomp}%
\usepackage{manyfoot}%
\usepackage{booktabs}%
\usepackage{algorithm}%
\usepackage{algorithmicx}%
\usepackage{algpseudocode}%
\usepackage{hyperref}
\usepackage{caption}

\usepackage[utf8]{inputenc}
\usepackage[T1]{fontenc}

\def\BibTeX{{\rm B\kern-.05em{\sc i\kern-.025em b}\kern-.08em
    T\kern-.1667em\lower.7ex\hbox{E}\kern-.125emX}}

\begin{document}

\title[RubiSCoT: A Framework for AI-Supported Academic Assessment]{RubiSCoT: A Framework for AI-Supported Academic Assessment}

%%=============================================================%%
%% Prefix	-> \pfx{Dr}
%% GivenName	-> \fnm{Joergen W.}
%% Particle	-> \spfx{van der} -> surname prefix
%% FamilyName	-> \sur{Ploeg}
%% Suffix	-> \sfx{IV}
%% NatureName	-> \tanm{Poet Laureate} -> Title after name
%% Degrees	-> \dgr{MSc, PhD}
%% \author*[1,2]{\pfx{Dr} \fnm{Joergen W.} \spfx{van der} \sur{Ploeg} \sfx{IV} \tanm{Poet Laureate} 
%%                 \dgr{MSc, PhD}}\email{iauthor@gmail.com}
%%=============================================================%%

\author{\fnm{Thorsten} \sur{Fröhlich}}\email{thorsten.froehlich@iu.org}

\author{\fnm{Tim} \sur{Schlippe}}\email{tim.schlippe@iu.org}

\affil{\orgdiv{IU International University of Applied Sciences}, \country{Germany}}

%%==================================%%
%% sample for unstructured abstract %%
%%==================================%%

\abstract{The evaluation of academic theses is a cornerstone of higher education, ensuring rigor and integrity. Traditional methods, though effective, are time-consuming and subject to evaluator variability. This paper presents \textit{RubiSCoT}, an AI-supported framework designed to enhance thesis evaluation from proposal to final submission. Using advanced natural language processing techniques, including \textit{large language models}, \textit{retrieval-augmented generation}, and \textit{structured chain-of-thought} prompting, \textit{RubiSCoT} offers a consistent, scalable solution. The framework includes preliminary assessments, multi-dimensional assessments, content extraction, rubric-based scoring, and detailed reporting. We present the design and implementation of \textit{RubiSCoT}, discussing its potential to optimize academic assessment processes through consistent, scalable, and transparent evaluation.}


\maketitle

\vspace{-0.5cm}

\section{Introduction}

The assessment of academic theses is crucial for maintaining educational standards and promoting scholarly development. Traditional evaluation relies on human expertise, demanding significant time and resources, which can lead to inconsistencies and biases~\cite{demszky2023automated}. The integration of artificial intelligence (AI) presents an opportunity to enhance and streamline academic assessments, ensuring greater uniformity and efficiency.

This paper introduces \textit{RubiSCoT} (Rubric-Based Structured Chain-of-Thought), an AI framework that leverages \textit{large language models} (LLMs), \textit{retrieval-augmented generation} (RAG), and \textit{structured chain-of-thought} (SCoT) prompting to provide systematic and comprehensive assessment from proposal to final submission. \textit{RubiSCoT} addresses key challenges in thesis evaluation: alignment with academic standards, detailed feedback, and scalability. By automating assessments and offering structured insights, \textit{RubiSCoT} improves accuracy and fosters a constructive feedback loop for students.

The next section reviews related work in AI-assisted academic assessments. Section~\ref{RubiSCoT} provides a detailed description of the \textit{RubiSCoT} framework and its implementation. In Section~\ref{Discussion}, we discuss the strengths and limitations of \textit{RubiSCoT}. Finally, Section~\ref{Conclusion and Future Work} concludes our work and outlines future directions.

\section{Related Work}
\label{Related Work}

This section reviews AI-supported academic assessment, covering automated grading, rubric-based evaluation, and LLM integration. It examines AI's historical development, challenges in grading, and efforts to enhance \textit{explainability}. The review highlights progress and gaps addressed by \textit{RubiSCoT}.

\subsection{Historical Development of AI in Education}

The integration of AI into educational settings has seen significant advancements over the decades. Early work by \cite{page1966imminence} in 1966 explored the potential for computers to grade essays, proposing automated essay scoring systems that evaluate written compositions based on pre-defined criteria like grammar, structure, and coherence. Subsequent research expanded on these concepts, demonstrating the increasing reliability and accuracy of such systems, a trend that has persisted into recent years~\cite{page1994new, Rudner_Liang_2002}. 
AI's role in education has further evolved into areas such as \textit{intelligent tutoring systems}~\cite{libbrecht2020nlp}, \textit{learning analytics}~\cite{bothmer2022investigating,bothmer2022skill,bothmer2023skill}, and \textit{gamification}~\cite{schlippe2021exam}, tools that help both educators and students enhance learning experiences. The emergence of generative AI, exemplified by GPT-3, demonstrated LLMs' potential in education~\cite{brown2020}.

\subsection{Automatic Short Answer Grading and Essay Scoring}

AI-based grading systems have evolved significantly since \cite{page1966imminence}'s early work on automated essay scoring in 1966. These systems evaluate written responses---ranging from short, free-text answers to full-length essays---based on criteria such as grammar, content, organization, and coherence~\cite{page1994new,ijcai2019p879,Misgna2024}. 
Recent advances in deep learning and multilingual modeling demonstrate that AI can grade responses accurately across different languages and question formats~\cite{sawatzki2021deep,schlippe2021cross}, making automated scoring a scalable solution in diverse educational settings. Research on explainability highlights that transparent scoring rationales---such as displaying predicted points, error matches, or intermediate reasoning---can increase user trust and enhance learning outcomes~\cite{schlippe2022explainability, hong2024my}. At the same time, achieving ``human-like'' grading remains challenging. Studies emphasize the need for adaptive feedback, especially as LLMs continue to improve~\cite{demszky2023automated,kusam2024generative}. \textit{Chain-of-thought} prompting, for example, has shown promise in refining LLMs’ grading logic ~\cite{cohn2024chain}, suggesting that more transparent and context-aware methods can further reduce bias and improve reliability.

% Automatic short answer grading uses AI to evaluate free-text responses, overcoming traditional grading limitations. Studies by \cite{ijcai2019p879} and \cite{Misgna2024} highlight the evolution and value of automated scoring. \cite{sawatzki2021deep} showed multilingual models can accurately grade responses across languages, addressing global educational challenges~\cite{schlippe2021cross}. Research on \textit{explainability} and \textit{transparency} shows that displaying predicted points and answer matches increases trust and improves learning outcomes~\cite{schlippe2022explainability, hong2024my}.

% \subsection{Automated Essay Scoring}

% Automated essay scoring has been widely studied since \cite{page1966imminence}'s work in 1966. AI evaluates essays based on grammar, content, organization, and coherence~\cite{page1994new}. Advances in deep learning, such as BERT, have improved scoring accuracy across NLP tasks~\cite{devlin2019}. Studies by \cite{demszky2023automated} and \cite{kusam2024generative} highlight AI’s role in feedback and assessment.  
% \cite{ijcai2019p879} and \cite{Misgna2024} discuss ongoing challenges in achieving human-like grading, emphasizing \textit{explainability} and adaptability. \textit{Chain-of-thought} prompting shows promise in enhancing LLMs' grading reasoning~\cite{cohn2024chain}.  

\subsection{Thesis and Dissertation Evaluation}

% Evaluating academic theses and dissertations is a critical process that traditionally relies on extensive manual review. \cite{hemmer2023art}'s ``The Art of Thesis Writing'' guides students through this process, emphasizing a structured approach to research and manuscript preparation. \cite{wu2025unveilingscoringprocessesdissecting}~studied the alignment gap between human and LLM grading \textit{rubrics}, underscoring the importance of high-quality analytic \textit{rubrics} for accurate scoring. Similarly, \cite{xie2024gradelikehumanrethinking}~proposed an LLM-based grading system that encompasses the entire grading procedure, enhancing accuracy and fairness through systematic rubric design and post-grading review.

Evaluating academic theses traditionally requires extensive manual review. \cite{hemmer2023art}'s ``The Art of Thesis Writing'' emphasizes a structured approach to research and manuscript preparation. \cite{wu2025unveilingscoringprocessesdissecting}~highlighted the alignment gap between human and LLM grading, stressing the need for high-quality analytic \textit{rubrics}. Similarly, \cite{xie2024gradelikehumanrethinking}~proposed an LLM-based grading system that improves accuracy and fairness through systematic rubric design and post-grading review.

\subsection{Rubric-Based Assessment}

\begin{figure}[h!]
  \centering
  \includegraphics[width=1.0\linewidth]{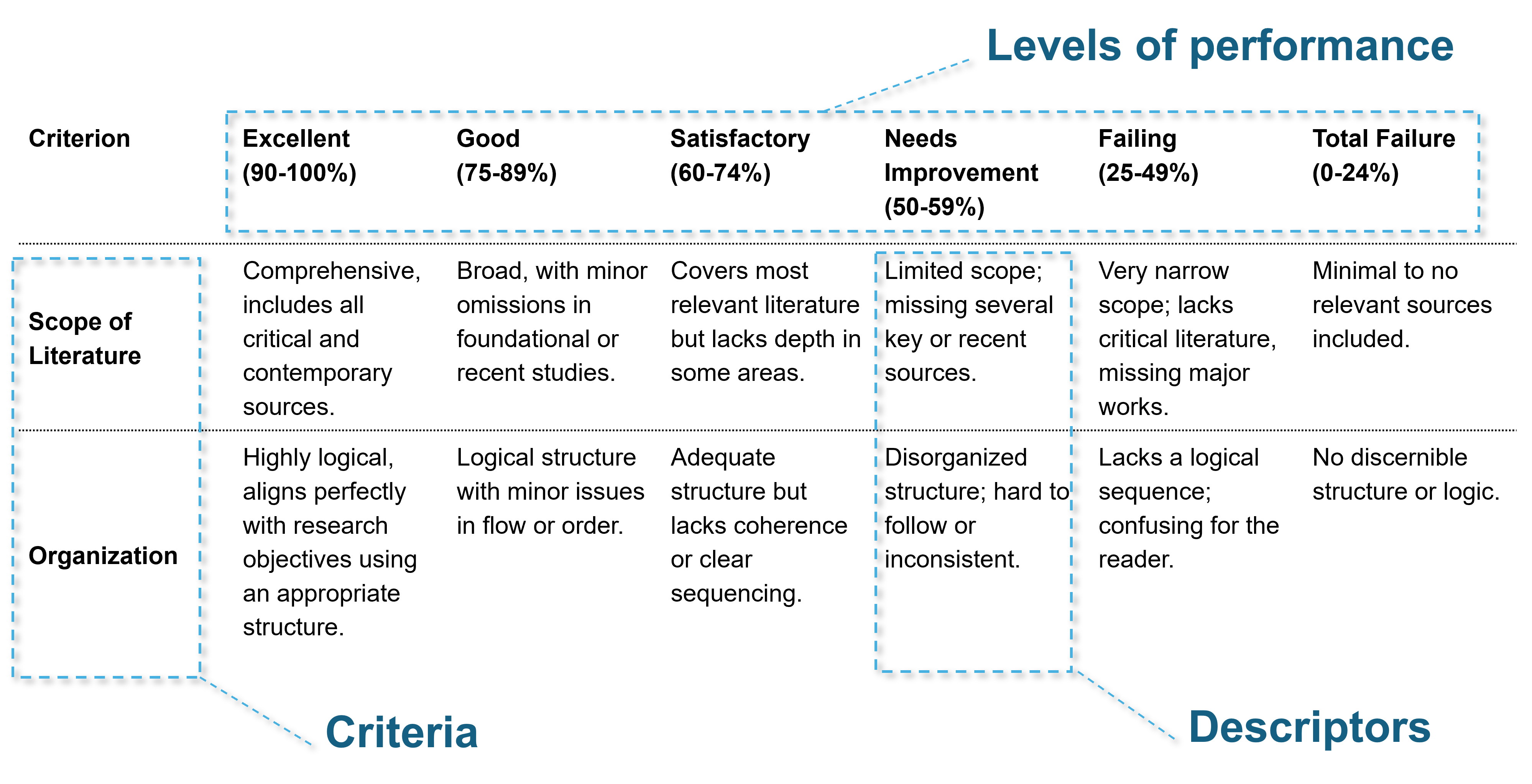}
  \caption{Example of Rubrics.}
  \label{fig:rubrics}
\end{figure}

As shown in Figure~\ref{fig:rubrics}, \textit{rubrics} are structured frameworks or scoring guides used to evaluate performance or outputs based on predefined criteria. They include three key components: \textit{criteria}, defining the aspects being assessed; \textit{levels of performance}, establishing quality gradations (e.g., Excellent, Good, Satisfactory); and \textit{descriptors}, specifying characteristics required at each level.

Three main types of rubrics are used: \textit{holistic}, \textit{single-point}, and \textit{analytic rubrics}~\cite{Glantz2024}. \textit{Holistic rubrics} give an overall judgment with a single score based on quality. \textit{Single-point rubrics} focus on expected standards for each criterion, allowing feedback on deviations. \textit{Analytic rubrics} break the evaluation into components for detailed assessments.

By clearly articulating expectations, \textit{rubrics} enhance consistency, objectivity, and transparency in assessment. They support both summative and formative evaluation, offering structured feedback that helps learners understand their strengths and areas for improvement. AI-based rubric assessments, as shown by \cite{johnson2024ai} and \cite{hashemi2024llm}, demonstrate AI's reliability in scoring. \cite{yavuz2025utilizing} explored LLMs and rubrics for grading foreign language essays, showing strong reliability and alignment with human raters.

\subsection{Retrieval-Augmented Generation}

\textit{Retrieval-Augmented Generation} (RAG) is an AI framework that enhances text generation by integrating external knowledge retrieval with a generative model, improving accuracy and reducing hallucinations. It is widely used in question answering, educational AI, and automated assessment systems~\cite{lewis2020,izacard-grave-2021-leveraging}. Studies like~\cite{jauhiainen2024evaluating} explored RAG's application in evaluating open-ended responses, showing its potential to improve grading quality and consistency. \cite{levonian2023retrieval} demonstrated RAG's effectiveness in math question-answering, illustrating how external knowledge enhances educational AI.

\subsection{Learning Analytics, Explainability, and Transparency}

\textit{Learning analytics} and \textit{explainability} are vital in educational assessment, ensuring AI systems are transparent and provide actionable feedback. Studies by \cite{johnson2024ai} and \cite{schlippe2022explainability} stress the need for transparency in AI-assisted grading, advocating for methods that clarify score generation and provide detailed student feedback. For example, Turnitin's AI writing detection model highlights the importance of identifying AI-generated content and maintaining academic integrity in submissions \cite{Turnitin2023,Bryson:2024}.

\subsection{Large Language Models}

The advent of LLMs has revolutionized AI in educational assessment. LLMs like GPT-3 and GPT-4 excel in NLP tasks such as grading, feedback generation, and rubric-based evaluations~\cite{brown2020, devlin2019}. Studies by \cite{wu2025unveilingscoringprocessesdissecting} and \cite{xie2024gradelikehumanrethinking} highlight LLMs' ability to improve grading accuracy and feedback, emphasizing the need for systematic approaches. \cite{impey2025using} underscores LLMs' reliability in grading science writing, suggesting their potential for scalable automated grading solutions.

\subsection{Chain-of-Thought Prompting}

Chain-of-thought prompting improves LLMs' reasoning by mimicking human problem-solving. \cite{cohn2024chain} applied this method to evaluate formative science assessments, combining few-shot and active learning to score short-answer responses effectively. Integrating \textit{structured chain-of-thought} prompting into educational AI systems like \textit{RubiSCoT} enhances evaluation depth and accuracy, offering meaningful explanations alongside scores.

\subsection{Research Gaps Addressed by RubiSCoT}

Despite advancements in AI-driven academic assessment, key gaps remain. Many models struggle with scalability, consistency, and rubric alignment, focusing mainly on essay and short-answer grading rather than full thesis evaluation. \textit{Transparency} and \textit{explainability} also limit educator trust and adoption. \textit{RubiSCoT} addresses these challenges with LLMs, RAG, and \textit{structured chain-of-thought} prompting, ensuring consistency, rubric alignment, and clear feedback. By balancing accuracy with transparency, \textit{RubiSCoT} offers a scalable, user-friendly solution for AI-assisted assessment.

\section{RubiSCoT for Academic Assessment}
\label{RubiSCoT}

In this section, we will present the key principles and components of \textit{RubiSCoT} for AI-supported academic assessment. The assessment process relies on documents provided via a RAG system or uploaded manually, including:
\begin{enumerate}
\item \textbf{Document to be evaluated} (e.g., master's thesis, bachelor's thesis, capstone project);
\item \textbf{Supporting documents} containing information about the assessment process.
\end{enumerate}

\subsection{Origins and Development of RubiSCoT}

The  \textit{RubiSCoT} framework was not conceived as a speculative concept but emerged through iterative empirical inquiry, grounded in practical evaluation needs and validated by field-based design logic. It follows the principles of Design Science Research (DSR) \cite{Simon1969}, which emphasizes artifact creation through cycles of problem identification, artifact design, demonstration, and evaluation. In this context, \textit{RubiSCoT} was developed in response to identifiable gaps in thesis assessment — namely, inconsistency, lack of transparency, and scalability issues.

Initial versions of \textit{RubiSCoT} were piloted with actual thesis documents across disciplines, using empirical case study analysis to refine prompt structures, rubric design, and evaluation flows. These iterations were informed by feedback from both human graders and institutional stakeholders. The framework's structure, including its rubric-based scoring and \textit{structured chain-of-thought} prompting, reflects inductive synthesis from practical trials rather than top-down theoretical abstraction.

By adopting both qualitative and design-based methodology, \textit{RubiSCoT} occupies the intersection between applied AI systems and academic pedagogy, offering an evaluative tool shaped by empirical constraints, not commercial intention. Its methodological grounding aligns with DSR logic, rubric research, and cognitive feedback studies, establishing its legitimacy as a research-based framework.

\subsection{RubiSCoT's Key Principles}

The following paragraphs outline key principles, including rubric-based evaluation, \textit{structured chain-of-thought} prompting, and retrieval-augmented generation, applied in each component of the \textit{RubiSCoT} framework to ensure effective and reliable evaluation processes.

\subsubsection{Rubric-Based Evaluation}

\textit{RubiSCoT} utilizes detailed \textit{rubrics} to assess each thesis section based on predefined criteria, providing clear and measurable assessments~\cite{anderson2001, hashemi2024llm,yavuz2025utilizing}. The rubric-based evaluation process is characterized by three key aspects:
\begin{itemize}
  \item \textbf{Systematic approach}: Ensures objective evaluations, highlighting strengths and areas for improvement.
  \item \textbf{Detailed feedback}: Focuses on both weaknesses and strengths, offering a comprehensive thesis overview. 
  \item \textbf{Time efficiency}: Streamlines evaluation, allowing quick identification of critical points to save time.
\end{itemize}

Creating rubrics involves several challenges: it can be time-consuming, must align with learning outcomes, and risks over-standardization. Solutions include using pre-existing literature, starting with templates, and involving stakeholders to ensure adaptability. Our rubric development followed a 5-step process: (1)~define objectives and outcomes, (2)~identify evaluation criteria, (3)~assign weights, (4) develop performance descriptors, and (5)~refine the rubric.

We used ChatGPT in Version GPT-4o to generate initial rubric elements based on \cite{hemmer2023art}, refining them to meet our needs. This automation ensures systematic evaluation by aligning responses with predefined criteria. LLMs like ChatGPT provide consistency, detailed feedback, and transparency in grading. The system is customizable and enhances efficiency by processing large volumes of data quickly, reducing thesis assessment time.

\subsubsection{Structured Chain-of-Thought Prompting}

To ensure ChatGPT follows a structured evaluation process, we use \textit{structured chain-of-thought} prompting. Unlike direct prompting, which assesses broadly and can miss details, \textit{structured chain-of-thought} breaks tasks into structured steps, addressing specific aspects of the task. By generating intermediate results, it provides context-rich insights that guide the final evaluation, enhancing precision and reducing ambiguity. Additionally, \textit{structured chain-of-thought} is adaptable, making it suitable for diverse domains like thesis evaluation. For example, it first assesses research question clarity, then evaluates their alignment with methodology to ensure thoroughness and coherence.

\subsubsection{Retrieval-Augmented Generation}

\textit{Retrieval-Augmented Generation} (RAG) integrates external knowledge sources, ensuring each evaluation aligns with the latest academic standards and domain-specific guidelines~\cite{lewis2020}. Beyond commonly referenced materials such as citation formatting rules, \textit{RubiSCoT} also leverages human-generated expectation documents, which may include detailed requirements for particular thesis sections (e.g., the research design). These documents, created by instructors or subject-matter experts, act as authoritative references specifying performance indicators, methodological standards, or content expectations.

By retrieving and consulting these expectation documents, \textit{RubiSCoT} provides more reliable and context-aware assessments. This approach mitigates over-reliance on the LLM’s internal knowledge and reduces the risk of hallucinations. Ultimately, the dual use of LLM-informed feedback and human-curated references ensures accurate, domain-relevant, and up-to-date evaluations of academic theses.

\subsection{RubiSCoT's Key Components}

\begin{figure}[h!]
  \centering
  \includegraphics[width=0.8\linewidth]{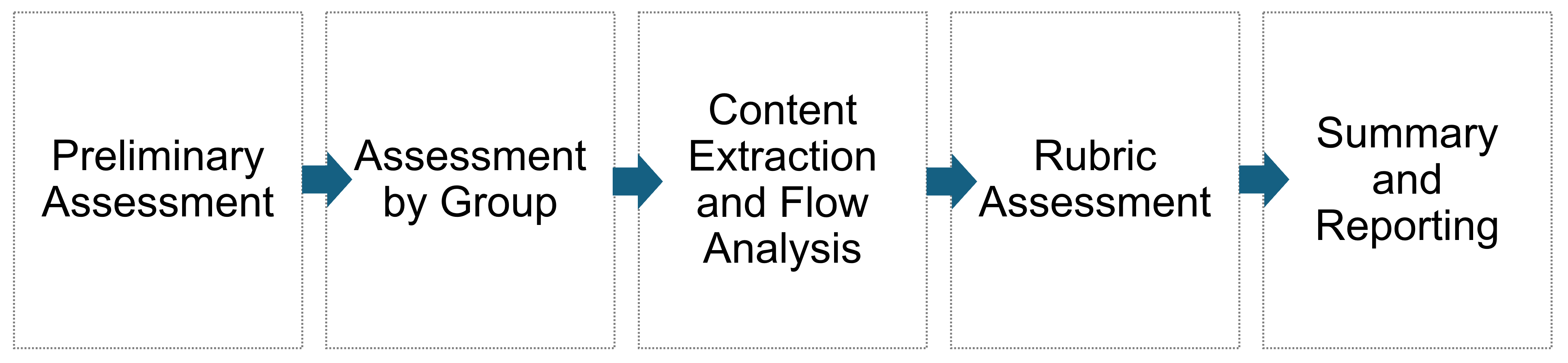}
  \caption{RubiSCoT’s Key Components.}
  \label{fig:components}
\end{figure}

As shown in Figure~\ref{fig:components}, \textit{RubiSCoT} consists of sequential components for thesis evaluation: \textit{preliminary assessment}, \textit{assessment by group}, \textit{content extraction and flow analysis}, \textit{rubric assessment}, and \textit{summary and reporting}. Each component uses prompts to guide the LLM, and to ensure consistency, we recommend configuring parameters to minimize variability, such as setting a low temperature to reduce randomness for stable, reproducible outputs.

In our implementation with ChatGPT, a \textit{base prompt} prioritized coherent and precise responses while minimizing creativity, which is less relevant for structured evaluations. The following \textit{base prompt} was effective:

\begin{lstlisting}[breakautoindent=false,breakindent=0ex,xleftmargin=0ex]
Please provide responses focusing on coherent and straightforward information where creativity is less valued.
\end{lstlisting}

While specific parameter controls may vary across LLM architectures, this approach ensures a structured and reproducible evaluation process, maintaining alignment with the intended assessment framework.

\subsubsection{Preliminary Assessment}

The \textit{preliminary assessment} component ensures a thesis aligns with the expected academic level, whether Bachelor's or Master's. As the first step in \textit{RubiSCoT}, it verifies structural and conceptual readiness, identifying issues such as missing research questions or unclear objectives. A Master's thesis is expected to show more knowledge transfer and innovation than a Bachelor's thesis, a distinction recognized in English-speaking academia. This component’s goal is to understand the thesis’s context and ensure proper academic alignment before further analysis. If fundamental elements are missing, the assessment halts, preventing premature evaluation. This step is also integrated into the supervision process.

Our implementation uses the International Standard Classification of Education (ISCED) \cite{unesco2012isced} for academic level alignment, though the framework is adaptable to other classification standards. The assessment is operationalized through a structured prompt to ensure the LLM correctly identifies the thesis level before applying level-specific criteria:

\begin{lstlisting}[breakautoindent=false,breakindent=0ex,xleftmargin=0ex]
Check the attached document.
If it says Bachelor Thesis go to BACHELOR, else go to MASTER.
\end{lstlisting}

For example, in the case of a Bachelor's thesis, the following structured assessment is applied:  

\begin{lstlisting}[breakautoindent=false,breakindent=0ex,xleftmargin=0ex] 
Review the attached bachelor's thesis, assessing its academic level based on the following criteria:
\end{lstlisting}
\vspace{-0.21cm}
\begin{lstlisting}
1.  Research Depth and Complexity:
1.1 Assess the depth of research, focusing on the integration of
    theoretical frameworks and practical applications as expected at
    the bachelor's level.  
1.2 Determine if the thesis demonstrates an understanding and application of concepts appropriate for undergraduate study.  
\end{lstlisting}
\vspace{-0.21cm}
\begin{lstlisting}
2.  Methodological Rigor:
2.1 Examine the research design and methodology, evaluating whether a coherent and sound approach is applied (e.g., primary vs. secondary 
    data use, analysis methods, or case studies).  
2.2 Evaluate if the methodology reflects a clear understanding of research techniques suitable for Level 6.  
\end{lstlisting}
\vspace{-0.21cm}
\begin{lstlisting}
3.  Critical Analysis and Synthesis:
3.1 Look for evidence of critical analysis, focusing on the ability to compare and contrast relevant approaches or methods within the field of study.  
3.2 Assess whether the thesis synthesizes information from various sources and discusses the limitations and challenges relevant to the topic.  
\end{lstlisting}
\vspace{-0.21cm}
\begin{lstlisting}
4.  Scholarly Communication and Structure:
    ...
\end{lstlisting}

This structured approach ensures a systematic preliminary assessment, maintaining alignment with academic expectations for different degree levels while allowing adaptability to alternative classification frameworks.

\subsubsection{Assessment by Group}

\begin{figure}[h!]
  \centering
  \includegraphics[width=0.9\linewidth]{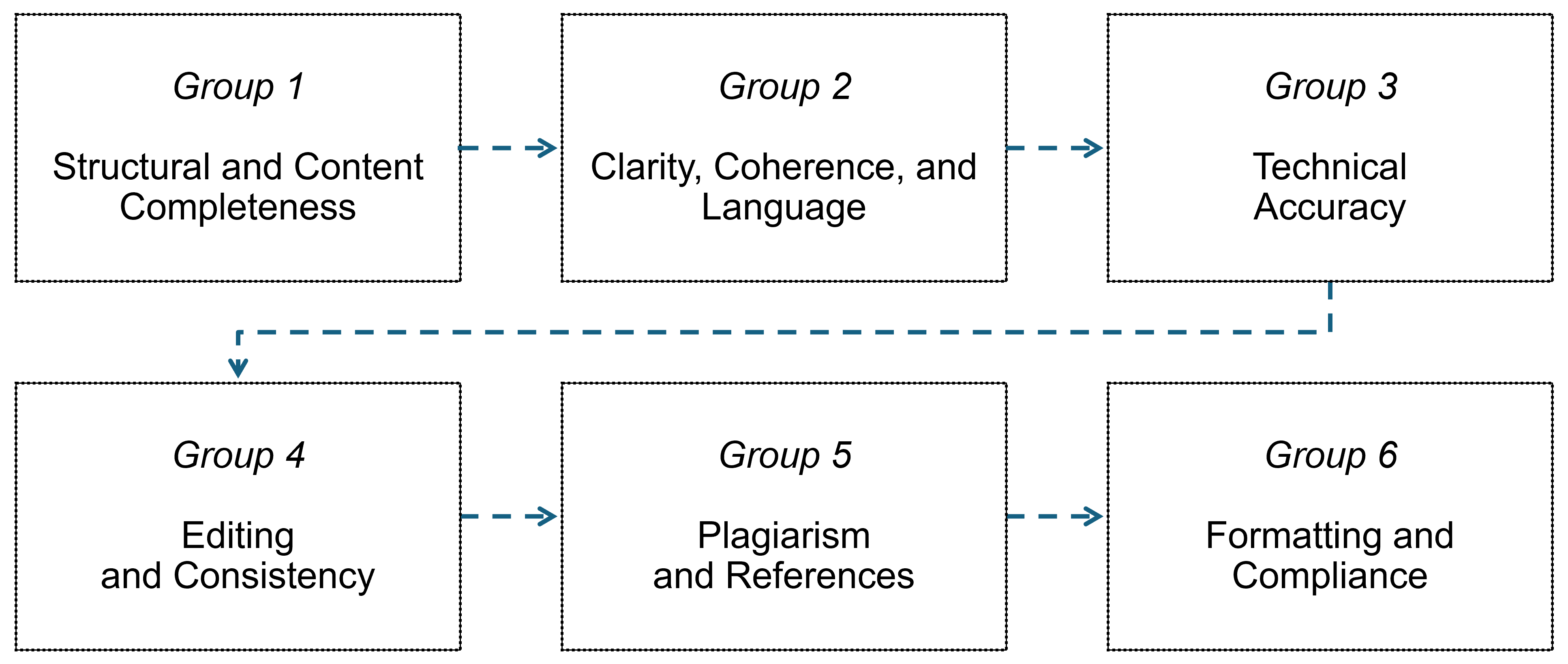}
  \caption{Assessment by Group.}
  \label{fig:reviewByGroup}
\end{figure}

As shown in Figure~\ref{fig:reviewByGroup}, \textit{RubiSCoT} performs a multi-dimensional evaluation in the \textit{assessment by group} component, divided into six groups: \textit{structural and content completeness}, \textit{clarity, coherence, and language}, \textit{technical accuracy}, \textit{editing and consistency}, \textit{plagiarism and references}, and \textit{formatting and compliance}. The evaluations follow a structured approach but remain flexible, allowing prompts to be applied as needed at any stage.

\paragraph{Structural and Content Completeness}  
This assessment group ensures essential sections (abstract, introduction, literature review, methodology, results, discussion, conclusion) are present and well-detailed. The evaluation is conducted through a structured prompt that systematically verifies the presence and adequacy of each section, ensuring logical flow and completeness:  

\begin{lstlisting}[breakautoindent=false,breakindent=0ex,xleftmargin=0ex]
Examine the thesis structure to ensure it includes all essential sections (e.g., abstract, introduction, literature review, methodology, 
results, discussion, conclusion). Evaluate whether each section is appropriately detailed and follows a logical sequence. Provide only observations and evidence. Do not give a summary. Remove any artefact from your answer.
\end{lstlisting}
\vspace{-0.21cm}
\begin{lstlisting}
Review the thesis to ensure the research questions or hypotheses are 
clearly stated and logically developed. Assess whether they are 
effectively addressed throughout the thesis. Provide only observations 
and evidence. Do not give a summary. Remove any artefact from your 
answer.
\end{lstlisting}
\vspace{-0.21cm}
\begin{lstlisting}
Analyze the strength of arguments in the thesis by assessing the 
supporting evidence provided. Highlight any arguments that may lack 
sufficient support. Provide only observations and evidence.
Do not give a summary. Remove any artefact from your answer.
\end{lstlisting}

This structured approach ensures a systematic assessment of the thesis structure, content completeness, and coherence, allowing for objective identification of missing or underdeveloped sections.

\paragraph{Clarity, Coherence, and Language}  
This assessment group assesses the logical flow and clarity of arguments, ensuring a formal academic tone. The evaluation uses a structured prompt to examine coherence, logical consistency, and adherence to academic language standards:

\begin{lstlisting}[breakautoindent=false,breakindent=0ex,xleftmargin=0ex]
Assess the thesis for clarity and coherence. Identify areas where arguments might be unclear or illogical. Remove any artefact from your answer.
\end{lstlisting}
\vspace{-0.21cm}
\begin{lstlisting}
Evaluate the thesis for appropriate language and tone. Ensure that the 
language used is formal and academic, avoiding colloquialisms or overly 
casual expressions. Provide feedback on sections where tone adjustment 
may be needed. Provide only observations and evidence. Do not give a 
summary. Remove any artefact from your answer.
\end{lstlisting}

This structured approach ensures that the thesis maintains logical clarity and a professional academic tone, facilitating clear communication of ideas.

\paragraph{Technical Accuracy}  
This assessment group examines the correctness of technical elements (equations, algorithms, terminologies), ensuring relevance and accuracy. The evaluation follows a structured prompt to assess these aspects:

\begin{lstlisting}[breakautoindent=false,breakindent=0ex,xleftmargin=0ex]
Evaluate the technical sections of the thesis (e.g., equations, algorithms, terminologies) for accuracy and relevance. Check that these 
elements are correctly applied. Provide only observations and evidence.
Do not give a summary. Remove any artefact from your answer.
\end{lstlisting}

This method ensures that technical components are rigorously reviewed for correctness and appropriate application within the thesis.

\paragraph{Editing and Consistency}  
This assessment group identifies grammatical errors and stylistic inconsistencies, ensuring adherence to academic writing standards. The evaluation is guided by a structured prompt designed to systematically assess these aspects:

\begin{lstlisting}[breakautoindent=false,breakindent=0ex,xleftmargin=0ex]
Analyze the thesis document for grammatical errors, awkward phrasing, and style inconsistencies. Highlight any sections needing revision. Provide only observations and evidence. Do not give a summary. Remove any artefact from your answer.
\end{lstlisting}
\vspace{-0.21cm}
\begin{lstlisting}
Conduct a consistency check across the thesis. Look for consistent use 
of terminology, figure and table numbering, abbreviations, and 
stylistic elements. List any inconsistencies. Provide only observations 
and evidence. Do not give a summary. Remove any artefact from your 
answer.
\end{lstlisting}

By applying this structured approach, the assessment ensures linguistic precision, clarity, and consistency throughout the thesis.

\paragraph{Plagiarism and References} 
This assessment group detects plagiarism and verifies citation accuracy to ensure academic integrity. The evaluation identifies potential unattributed text, checks textual similarities, and ensures correct citation and formatting. The following prompt guides this process:

\begin{lstlisting}[breakautoindent=false,breakindent=0ex,xleftmargin=0ex]
Please review the thesis for any sections that might contain unattributed text or potential similarities to existing works. Provide a list of these sections with an explanation of the potential issues. Provide only observations and evidences. Do not give a summary. Remove any artefact from your answer.
\end{lstlisting}
\vspace{-0.21cm}
\begin{lstlisting}
Perform a reference check on the thesis. Identify any inconsistencies 
in citation formats and verify that all references listed in the 
bibliography are cited in the text. Note any missing citations or 
discrepancies in format. Provide only observations and evidences. Do 
not give a summary. Remove any artefact from your answer. Perform a 
recheck of missing references.
\end{lstlisting}

\paragraph{Formatting and Compliance} 
This assessment group ensures adherence to formatting guidelines, including margins, font size, headings, and institutional requirements. It verifies the thesis meets academic standards for structure, presentation, and consistency. The following prompt is used for this assessment:

\begin{lstlisting}[breakautoindent=false,breakindent=0ex,xleftmargin=0ex]
Check the thesis document for adherence to formatting guidelines, such as margins, font size, heading styles, and spacing. Identify any deviations from the guidelines. Do not give a summary. Remove any artefact from your answer.
\end{lstlisting}
\vspace{-0.21cm}
\begin{lstlisting}
Check the thesis for compliance with academic standards or guidelines 
you provided. Ensure that the thesis meets these standards across 
content, structure, and presentation. Identify any areas that may not 
align with the standards. Provide only observations and evidences. Do 
not give a summary. Remove any artefact from your answer.
\end{lstlisting}

\subsubsection{Content Extraction and Flow Analysis}

\begin{figure}[h!]
  \centering
  \includegraphics[width=1.0\linewidth]{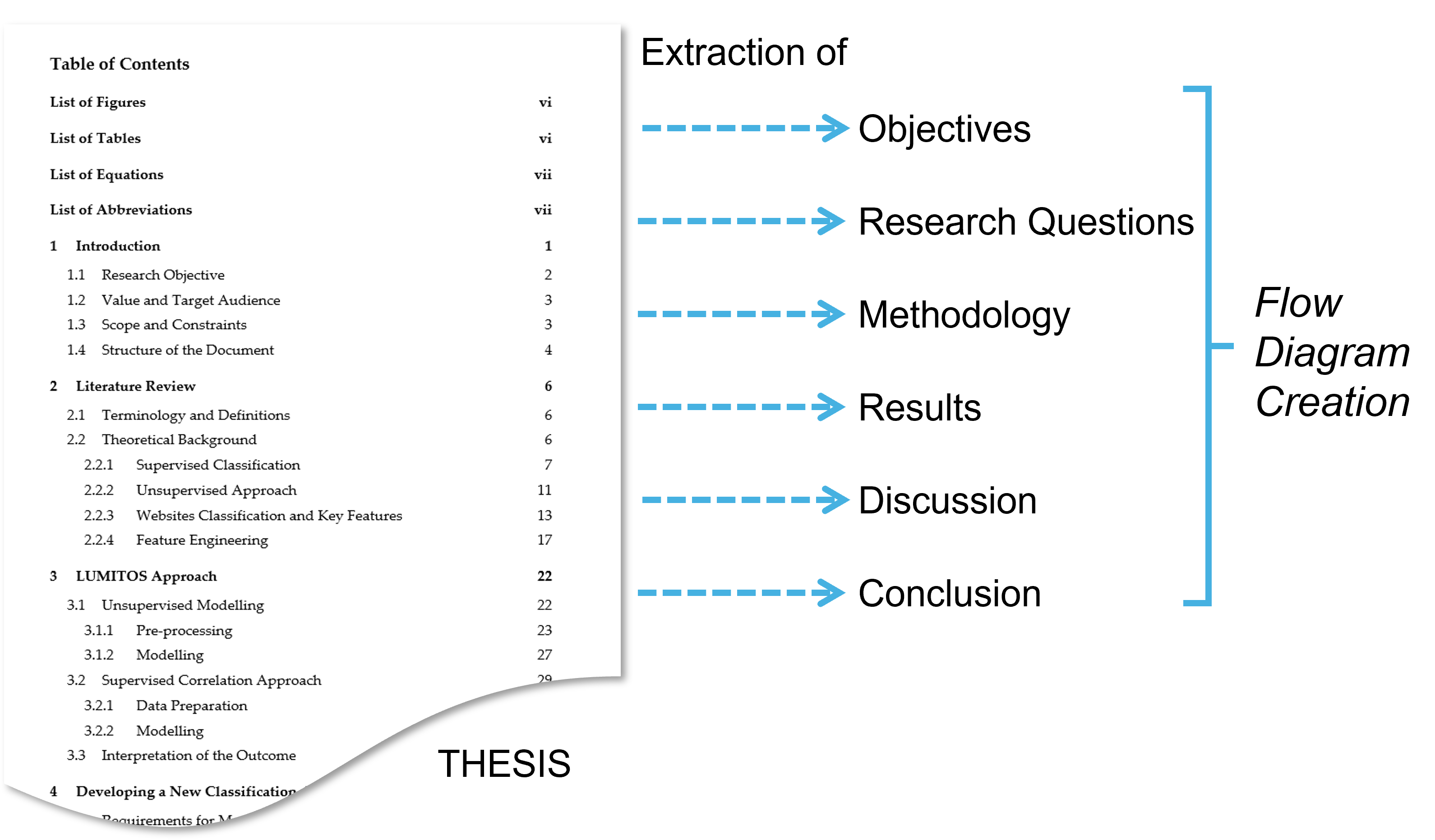}
  \caption{Content Extraction.}
  \label{fig:ContentExtraction}
\end{figure}

As shown in Figure~\ref{fig:ContentExtraction}, the \textit{content extraction and flow analysis} component identifies key elements like research questions, objectives, and methodologies, generating flow diagrams to map the thesis’s logical progression and coherence. It highlights how sections connect and align with research goals, focusing on the flow from objectives to conclusions.

\begin{figure}[h!]
    \centering
    \begin{subfigure}[b]{0.65\textwidth}
        \includegraphics[width=\linewidth]{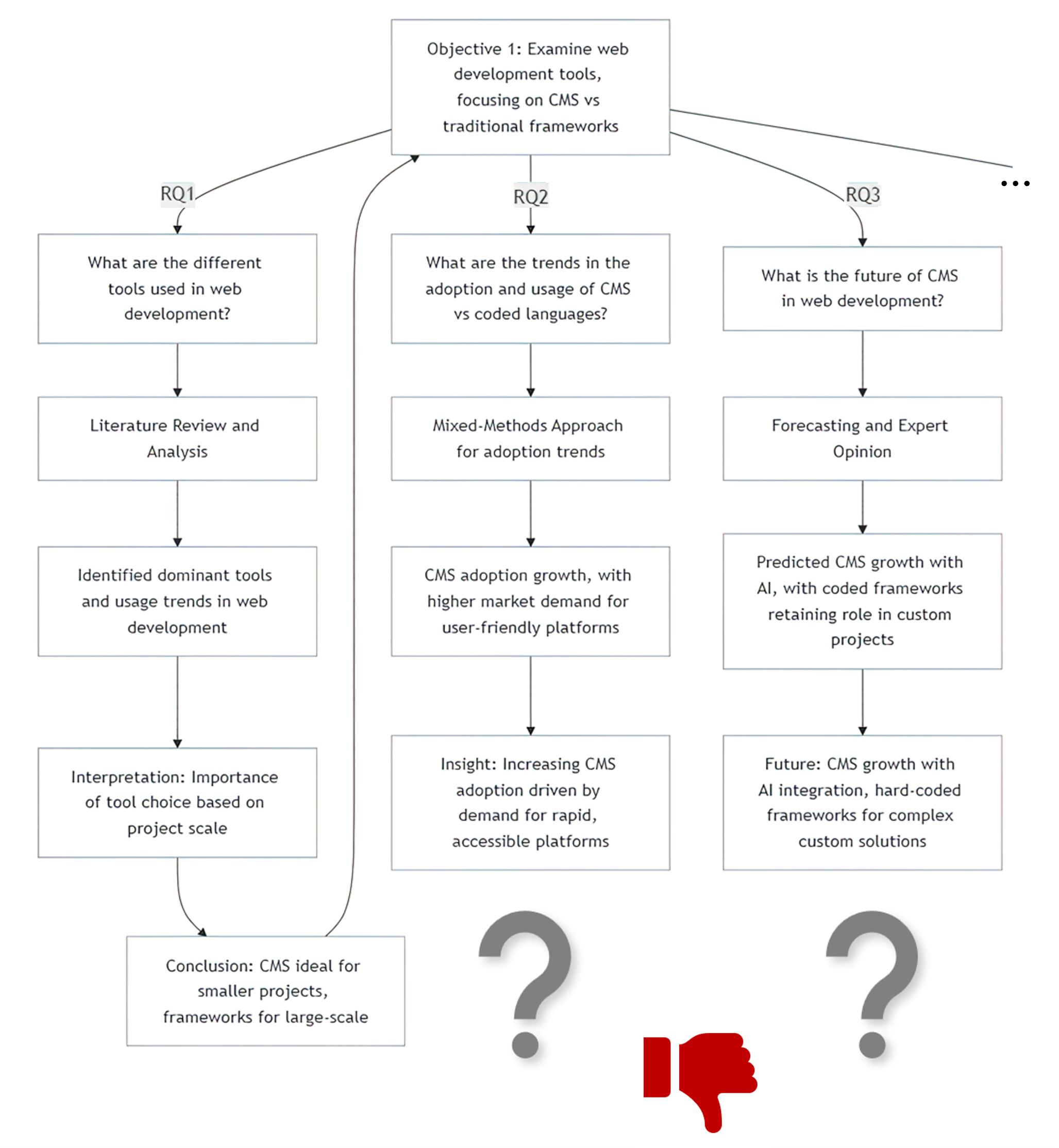} 
        \label{fig:FlowAnalysis_neg}
        \vspace{-0.65cm}
                \caption{Structural Weaknesses.}
    \end{subfigure}\\
            \vspace{0.2cm}
    \begin{subfigure}[b]{0.75\textwidth}
        \includegraphics[width=\linewidth]{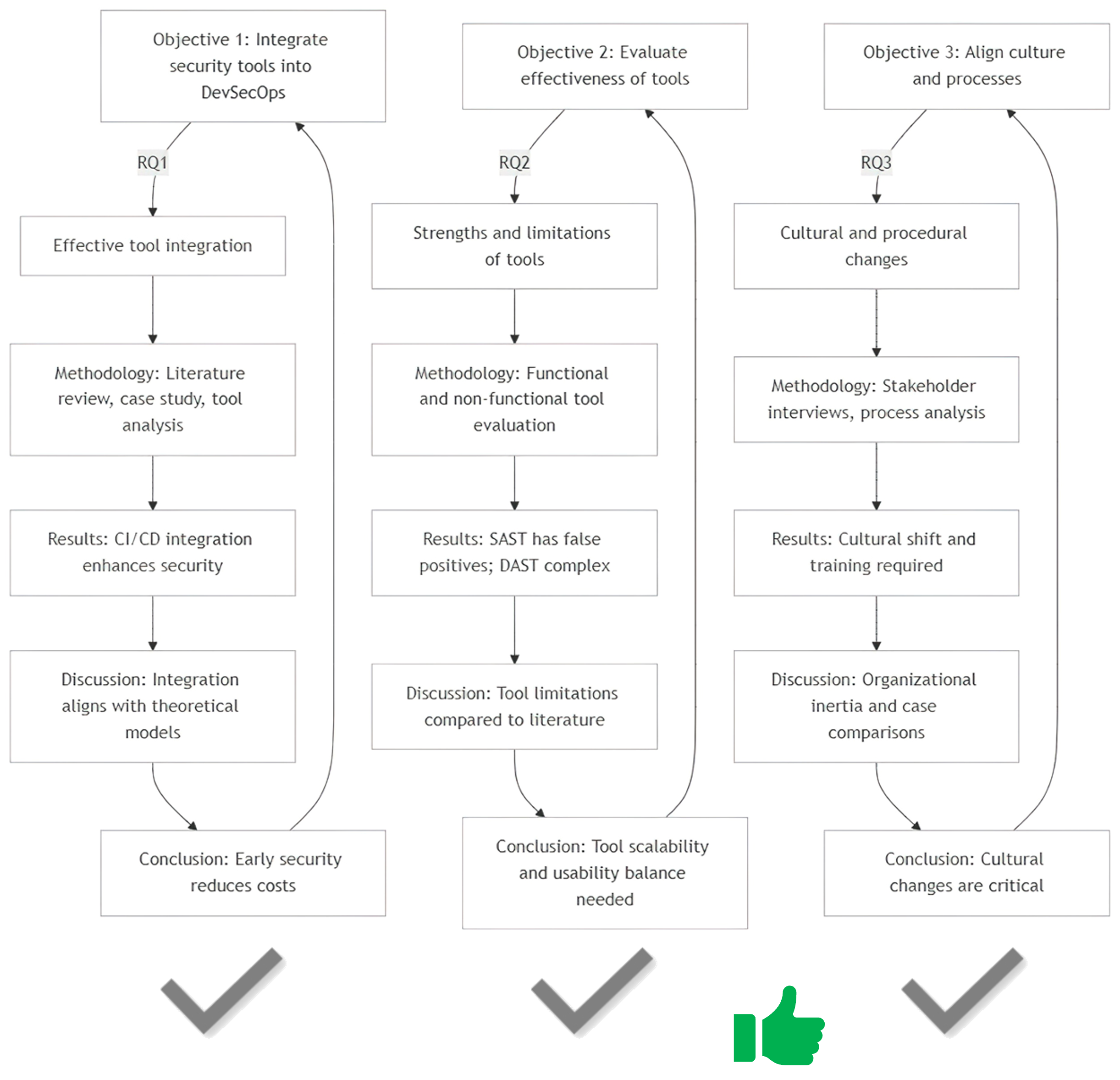}
        \label{fig:FlowAnalysis_pos}
        \vspace{-0.7cm}
        \caption{Ideal Structure.}
    \end{subfigure}
    \caption{Visual Flow Diagrams of the Thesis’s Logical Progression.}
    \label{fig:FlowAnalysis}
\end{figure}

To create these diagrams, our implementation uses \textit{Mermaid}\footnote{https://github.com/mermaid-js/mermaid}, a tool for structured flow visualization. Large language models generate the syntax for this tool, producing a visual representation of the thesis structure. This helps identify gaps in logical flow, such as weak connections between research questions, objectives, or conclusions.

As shown in Figure~\ref{fig:FlowAnalysis}(a), one visualization may highlight gaps, indicating structural weaknesses or misinterpretations. These gaps do not necessarily imply a poor thesis but point to areas needing refinement. In contrast, Figure~\ref{fig:FlowAnalysis}(b) illustrates an ideal structure with seamless alignment from objectives to conclusions, ensuring logical coherence and rigor. This analysis helps assess overall thesis integrity and guide improvements. 
The following prompt is used to conduct content extraction and flow analysis:

\begin{lstlisting}
Analyze the attached thesis carefully and extract the following
information:
\end{lstlisting}
\vspace{-0.21cm}
\begin{lstlisting}
- Objectives: Identify each main objective of the thesis as stated in 
  the introduction.
- Research Questions: List all research questions that the thesis seeks 
  to answer. For each research question, note the objective(s) it 
  addresses.
- Methodology: For each research question, summarize the methods used 
  to gather and analyze data. Include details on why these methods 
  were chosen, as provided in the text.
- Results: For each research question, summarize the main findings /
  results. Identify any figures, tables, or data representations linked 
  to these findings.
- Discussion: Briefly summarize how the thesis discusses the results in 
  relation to each objective. Note any major interpretations, 
  comparisons with literature, or critical insights.
- Conclusion: Outline the conclusions, recommendations, and 
  implications provided in the thesis. Link these conclusions to 
  the relevant objectives and research questions.
\end{lstlisting}
\vspace{-0.21cm}
\begin{lstlisting}
Format:
- Use bullet points or a table to organize each section.
- Include specific page or section references wherever possible.
- For each item, specify the objective(s) it supports.
\end{lstlisting}

\begin{figure}[h!]
  \centering
  \includegraphics[width=1.0\linewidth]{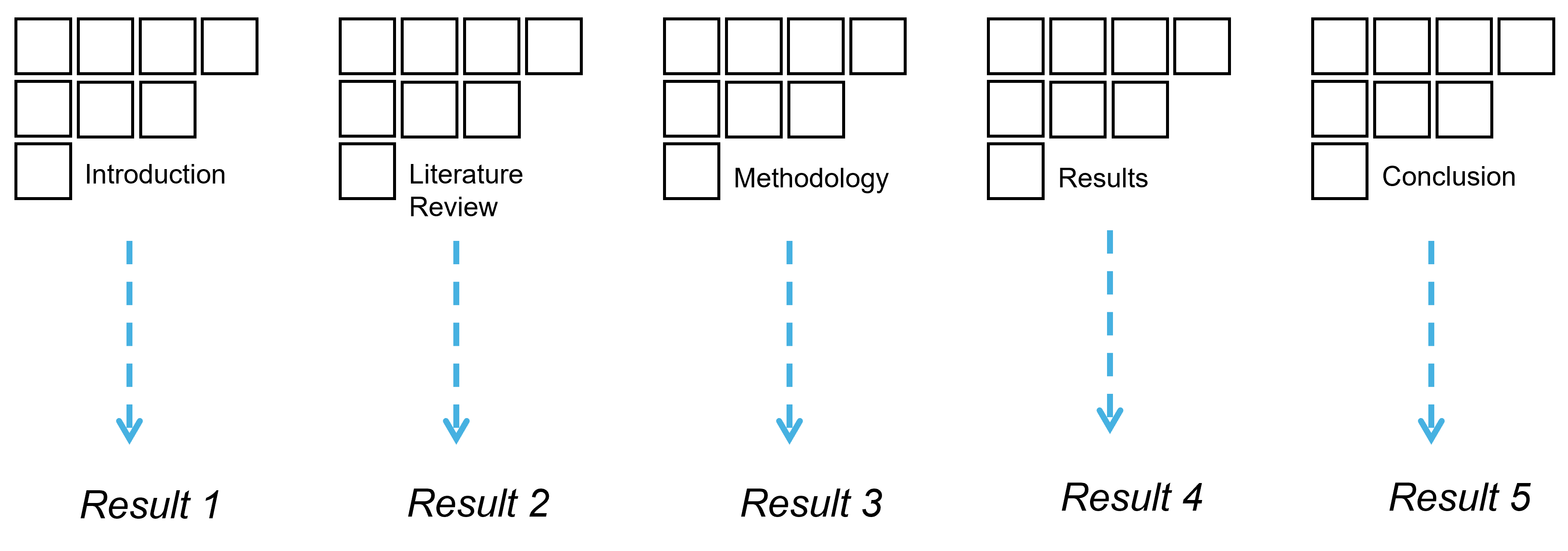}
  \caption{Rubric Assessment Process.}
  \label{fig:RubricReview}
\end{figure}

\subsubsection{Rubric Assessment}

As visualized in Figure~\ref{fig:RubricReview}, the \textit{rubric assessment} component provides percentage scores and qualitative feedback for each thesis section, highlighting strengths and areas for improvement. The evaluation follows a structured order, starting with the \textit{introduction}, followed by \textit{literature review}, \textit{methodology}, \textit{results}, and \textit{conclusion}. This ensures a comprehensive assessment, capturing key aspects of thesis quality across all major sections.

The following prompt guides the rubric creation for a specific chapter. 
The rubric follows established thesis writing principles, such as those in Chapter~17 of \cite{hemmer2023art}, but can be adjusted to meet specific evaluation needs.  

\begin{lstlisting}[breakautoindent=false,breakindent=0ex,xleftmargin=0ex]
Placeholder = 17 The Conclusion
1. Please provide a comprehensive summary of Chapter [Placeholder] from 
   the book on thesis writing. Focus on the main objectives, key 
   principles, and any specific advice or guidelines for thesis writing 
   covered in this chapter.
2. Based on the summary of Chapter [Placeholder], list the key 
   expectations or standards that a thesis should meet according to 
   this chapter. Outline these expectations clearly, emphasizing the 
   specific skills, structure, or content that a thesis section should 
   demonstrate if it follows this chapter's guidance.
3. Using the expectations from Chapter [Placeholder], create a rubric 
   with six performance levels to evaluate a thesis section according 
   to this chapter's principles. The levels should include:
   - Excellent (90-100%)  
   - Good (75-89%)  
   - Satisfactory (60-74%)  
   - Needs Improvement (50-59%)  
   - Failing (25-49%)  
   - Total Failure (0-24%)
   For each criterion, provide a detailed description of what is 
   required to achieve each performance level. Ensure that the 
   descriptions for 'Failing' and 'Total Failure' clarify the key 
   deficiencies compared to higher levels. The rubric should make each 
   expectation measurable and actionable, aligning closely with the 
   chapter's guidelines.
4. Review the rubric created for Chapter [Placeholder] and suggest any 
   adjustments or improvements that would make it more precise or 
   aligned with the chapter's standards. Ensure that each criterion is 
   actionable and that the rubric covers all key expectations from the 
   chapter.
5. Show the revised rubric for Chapter [Placeholder] as a table.
6. Create a word document from the rubric for Chapter [Placeholder].
\end{lstlisting}

Once the rubric is developed, it is applied systematically for thesis evaluation. The following prompt guides the rubric assessment process:

\begin{lstlisting}[breakautoindent=false,breakindent=0ex,xleftmargin=0ex]
Content check against rubrics (If the chapter names do not match, use a 
similar chapter in the thesis and indicate the deviation):  
1. Check the chapter Introduction of the attached thesis against the 
   Introduction Expectations document and the rubric Introduction and 
   give an evaluation and a reasoning in a table. Take the percentage 
   of each line of your evaluation and show them in a table and 
   calculate the average.
   ...
5. Check the chapter Conclusion of the attached thesis against 
   the Conclusion Expectations document and the rubric 
   Conclusion and give an evaluation and a reasoning in a table. 
   Take the percentage of each line of your evaluation and show them in 
   a table and calculate the average.
\end{lstlisting}
\vspace{-0.21cm}
\begin{lstlisting}
I want you to critically check it a second time under the same 
conditions to see if you stick with your first evaluation. Build an 
average of the original and revised results as a table.
\end{lstlisting}

This structured approach ensures that the rubric is both systematically developed and rigorously applied to evaluate thesis quality.

\subsubsection{Summary and Reporting}

Upon evaluation completion, \textit{RubiSCoT} generates comprehensive reports summarizing results, highlighting strengths, weaknesses, and areas for improvement. These reports offer actionable feedback, facilitating continuous academic writing enhancement as recommended by~\cite{demszky2023automated}. The transparent reporting builds trust in the evaluation outcomes among students and academic stakeholders.

\begin{figure}[h!]
  \centering
  \includegraphics[width=0.9\linewidth]{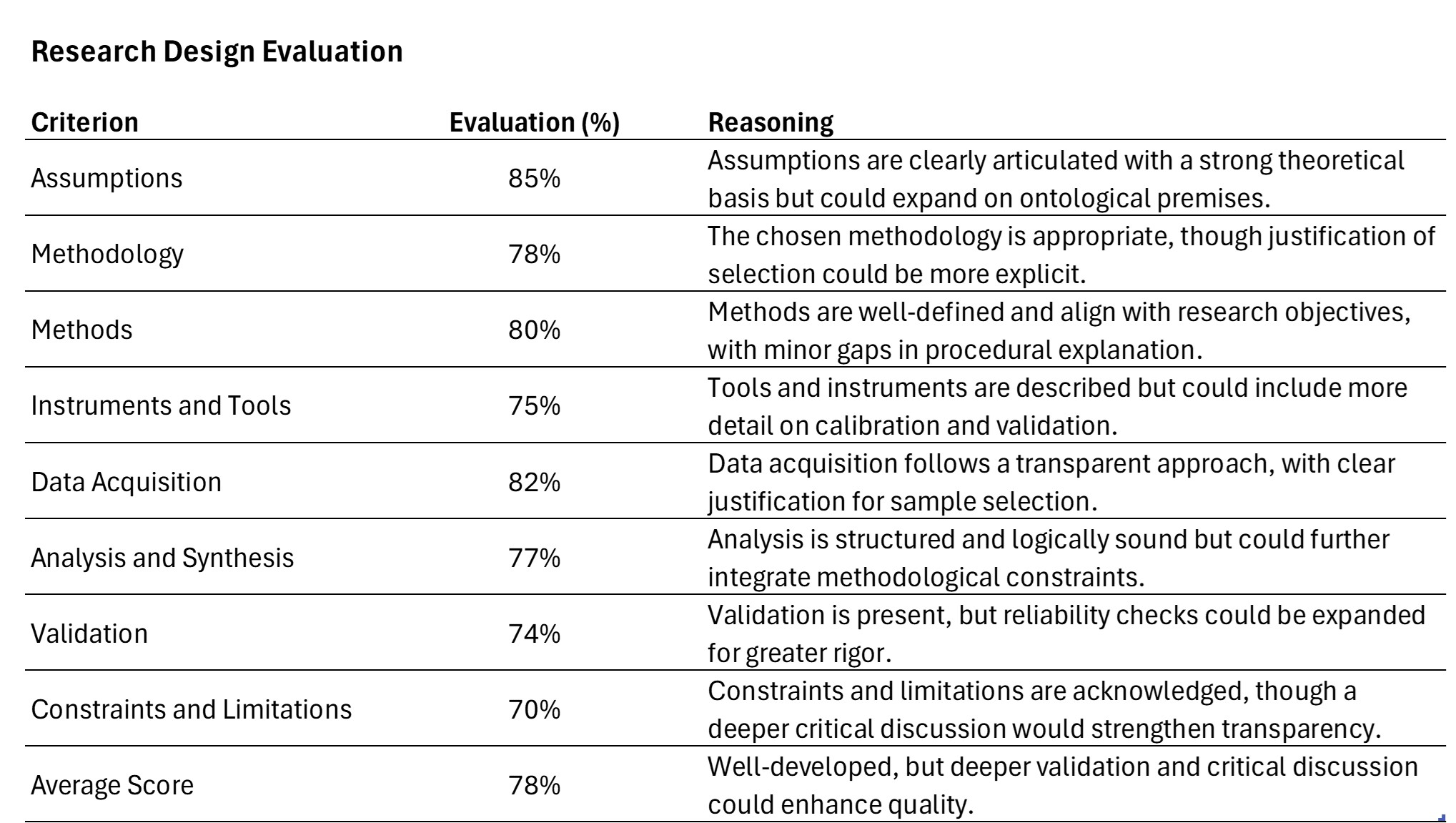}
  \caption{Methodology Evaluation Report.}
  \label{fig:report}
\end{figure}

Figure~\ref{fig:report} displays an example of the \textit{rubric assessment} component for the \textit{Methodology Evaluation} report. It illustrates how each thesis section is assessed along rubric-based criteria, with scores and explanatory remarks presented in tabular form. This standardized layout ensures transparent, systematic feedback, highlighting specific strengths and weaknesses to guide students’ revision and improvement.

\section{Discussion}
\label{Discussion}

This section discusses the strengths and limitations of \textit{RubiSCoT}, focusing on trust, reliability, scalability, efficiency, explainability, feedback quality, domain adaptability, and ethical considerations. %Insights from the implementation highlight its potential to enhance academic assessment in higher education.

\subsection{Trust and Reliability in AI Evaluations}
The integration of AI in assessments raises concerns about trust and reliability. While tools like \textit{ChatGPT} offer speed and consistency, final decisions should remain with human evaluators to prevent biases and uphold academic integrity. Educators value AI’s structured support but emphasize the need for human oversight in grading \cite{johnson2024ai, demszky2023automated}.

\subsection{Scalability and Efficiency}

\textit{RubiSCoT} is designed to address scalability challenges in thesis assessment. Traditional methods often suffer from judgment variability and are time-consuming \cite{demszky2023automated}. In theory, \textit{RubiSCoT}'s automated preliminary assessments and rubric-based scoring should provide more consistent evaluation processes, potentially offering timely feedback while reducing faculty workload. The structured nature of the framework suggests it could scale effectively in large institutional settings, though comprehensive empirical validation of these efficiency benefits remains an area for future research.

\subsection{Explainability and Feedback Quality}

A key aspect of AI-supported assessment is transparency and clarity in feedback. Studies show that detailed, understandable feedback enhances student learning and trust in evaluations \cite{hong2024my, schlippe2022explainability}. \textit{RubiSCoT} uses \textit{structured chain-of-thought} prompting to provide actionable, contestable feedback, helping students understand strengths and areas for improvement. This approach addresses critiques of AI as ``black boxes''~\cite{Rudin2019}, promoting a more transparent and interactive assessment process that distinguishes it from many AI-based evaluation systems.

First, the \textit{structured chain-of-thought} prompting technique makes the assessment process more interpretable by exposing the intermediate reasoning steps. Unlike typical neural network approaches that offer only final outputs, \textit{RubiSCoT}'s approach reveals how the system progresses from textual analysis to evaluation, providing both students and educators visibility into the reasoning process \cite{Wei2022}\cite{wiegreffe2021teach}. This aligns with recent developments in explainable AI (XAI), but applies these principles specifically to educational assessment contexts.

Second, \textit{RubiSCoT}'s use of explicit rubrics creates a transparent framework against which assessments are made. This differs significantly from fully automated approaches that may apply hidden or implicit criteria \cite{hashemi2024llm}. By making evaluation criteria explicit, the system enables contestability—students and instructors can directly challenge specific rubric applications rather than questioning an opaque judgment.

Third, the integration of human-created expectation documents through RAG ensures assessments remain anchored to institutionally-approved standards. This hybrid approach balances AI capabilities with human authority, addressing concerns about AI systems replacing human judgment rather than augmenting it \cite{Voessing2022}.

These transparency features address key limitations identified in previous AI assessment systems \cite{hussein2019automated}. However, challenges remain. The quality of explanations depends significantly on prompt engineering quality, and transparency can sometimes compete with efficiency goals. Furthermore, even with explicit reasoning chains, the underlying neural mechanisms of LLMs remain somewhat opaque. Future work should focus on further formalizing the connection between prompting strategies and explanation quality in assessment contexts.

\subsection{Domain-Specific Adaptability}

\textit{RubiSCoT}'s design principles suggest potential adaptability across different academic disciplines. The framework's reliance on configurable rubrics and structured prompting approaches provides a theoretical foundation for cross-disciplinary application. Recent advances in multilingual capabilities of LLMs~\cite{schlippe2021cross, agyemang2024ai} and dynamic rubric development methodologies~\cite{hemmer2023art, hashemi2024llm} indicate that systems like \textit{RubiSCoT} could be configured to meet diverse educational domains. While discipline-specific implementations would require tailored rubrics and domain knowledge, the underlying framework architecture is designed with this flexibility in mind. Future studies will need to explore how effectively this adaptability functions across various academic fields and institutional contexts.

\subsection{Ethical Considerations and Human Involvement}
The ethical use of AI in assessments is crucial. While AI tools provide substantial support, definitive grading should remain with human evaluators to ensure fairness and academic integrity. Using \textit{RubiSCoT} as a supportive tool, rather than a standalone evaluator, leverages AI's strengths in data processing and consistency while preserving human judgment \cite{schlippe2023artificial, Bryson:2024}.

While \textit{RubiSCoT} offers potential benefits for academic assessment, several important limitations must be acknowledged. First, the framework's effectiveness may vary based on the quality of rubrics and prompt engineering — poorly designed rubrics could lead to misleading or unhelpful assessments. Second, current LLM capabilities have inherent limitations in understanding complex disciplinary nuances that human experts can readily discern.

A critical concern is the risk of over-reliance on AI-generated assessments. \textit{RubiSCoT} is designed as a supportive tool rather than a replacement for human judgment, yet institutional pressures for efficiency might incentivize inappropriate delegation of evaluation responsibility to automated systems. To mitigate this risk, we recommend implementing \textit{RubiSCoT} within a human-in-the-loop workflow where faculty maintain final decision authority while using the system to enhance consistency and reduce cognitive load on routine assessment aspects.

Additionally, there are equity considerations regarding how AI-supported assessment may impact different student populations. Future research should investigate whether automated evaluation displays any systematic biases across demographic groups or writing styles. Transparent guidelines for appropriate use of AI-supported assessment tools are essential to maintain academic integrity and ensure fair treatment of all students.

Finally, as with any AI application in education, data privacy and security concerns must be addressed through appropriate institutional policies governing thesis data handling and LLM access patterns.

\section{Conclusion and Future Work}
\label{Conclusion and Future Work}

This section summarizes the key contributions of \textit{RubiSCoT} and outlines potential areas for future development. 

\subsection{Conclusion}

The development and implementation of \textit{RubiSCoT} represent significant advancements in AI-supported academic assessment. By leveraging LLMs, \textit{retrieval-augmented generation}, and \textit{structured chain-of-thought} prompting, \textit{RubiSCoT} provides a comprehensive, scalable solution for thesis evaluation. This framework addresses challenges such as alignment with academic standards, feedback provision, scalability, and transparency.

\textit{RubiSCoT} enhances thesis assessment by offering a structured approach that mitigates variability and supports consistent, high-quality feedback. Its integration of advanced AI techniques fosters a more efficient and transparent evaluation process, benefiting both educators and students. Its adaptability ensures relevance across diverse academic settings.

This paper focuses primarily on \textit{RubiSCoT}'s design principles and implementation details rather than extensive empirical validation. Initial testing within our institutional context shows promising results, with preliminary assessments indicating improvements in evaluation consistency and time efficiency. A comprehensive empirical evaluation across multiple institutions and disciplines is currently underway and will be reported in future work. This forthcoming research will quantify \textit{RubiSCoT}'s impact on assessment efficiency, consistency, and quality of feedback, providing empirical support for the scalability and adaptability claims discussed in this paper.

\subsection{Future Work}
While \textit{RubiSCoT} addresses many challenges in academic assessment, several avenues for future enhancement remain:

\begin{itemize} 
\item \textbf{Expansion of Multilingual Capabilities:} Extending support to more languages will increase inclusivity and effectiveness, allowing a broader range of institutions to adopt \textit{RubiSCoT}. 
\item \textbf{Refinement of Explainability Mechanisms:} Developing more sophisticated \textit{structured chain-of-thought} techniques will improve feedback interpretability and foster greater trust among educators, students, and administrators. 
\item \textbf{Integration with Learning Management Systems (LMS):} Seamless integration with institutional LMS platforms will facilitate \textit{RubiSCoT}’s widespread adoption and enable more direct incorporation into existing course workflows. 
\item \textbf{Enhanced Ethical Framework:} Strengthening ethical guidelines will ensure human judgment remains central to evaluations, promoting fairness, transparency, and academic integrity in automated assessments. 
\item \textbf{Broadening Application Scope:} Expanding \textit{RubiSCoT}’s usage beyond thesis evaluations (e.g., project-based coursework, peer-review processes) will support a broader range of educational activities.
\item \textbf{Field Validations:} To refine the system further and assess real-world effectiveness, we plan pilot studies with professors and students. These evaluations will help us gather direct feedback on usability, trust, and instructional impact, guiding iterative improvements to \textit{RubiSCoT}’s functionalities and user experience.
\end{itemize}

In conclusion, \textit{RubiSCoT} provides a promising pathway to more efficient, reliable, and transparent academic assessments, with ongoing opportunities for improvement. By balancing AI strengths with human oversight and integrating direct feedback from real users, \textit{RubiSCoT} continues to set a new standard for academic evaluation.

% \section*{Acknowledgments}

% This research was supported by the IU International University of Applied Sciences (\textit{IU Incubator}) under the internal funding framework for the period from October 2025 to September 2027.

\bibliography{00_main}% common bib file
%% if required, the content of .bbl file can be included here once bbl is generated
%%\input sn-article.bbl

%% Default %%
%%\input sn-sample-bib.tex%

\end{document}